\documentclass[conference]{IEEEtran}
\IEEEoverridecommandlockouts

\usepackage{amsmath,amssymb,amsfonts}
\usepackage{algorithmic}

\usepackage{textcomp}
\usepackage{xcolor}

\usepackage{times}  
\usepackage{helvet}  
\usepackage{courier}  
\usepackage[hyphens]{url}  
\usepackage{graphicx} 
\usepackage{natbib}  
\setcitestyle{numbers,square,comma,sort&compress}

\usepackage{caption} 
\usepackage{algorithm}
\usepackage{algorithmic}
\usepackage{amsmath}
\usepackage{amsfonts}
\usepackage{amssymb}

\usepackage{booktabs}
\usepackage{multirow}
\usepackage{newfloat}
\usepackage{listings}
\usepackage{hyperref}
\def\BibTeX{{\rm B\kern-.05em{\sc i\kern-.025em b}\kern-.08em
    T\kern-.1667em\lower.7ex\hbox{E}\kern-.125emX}}
\begin{document}

\title{Treatment Response Optimized Clinical Decision Support AI System via Digital Twin Simulation}

\author{%
\IEEEauthorblockN{Xinyu Qin$^{\dagger}$, Anil K. Sood$^{\ddagger}$, Ruiheng Yu$^{\dagger}$, Sara Corvigno$^{\ddagger}$, Elaine Stur$^{\ddagger}$, and Lu Wang$^{\dagger,\dagger\dagger,*}$}

\IEEEauthorblockA{%
$^{\dagger}$\textit{Department of Biomedical Engineering, University of Houston}, USA\\
\texttt{\{xqin5, ryu11\}@cougarnet.uh.edu, lwang71@central.uh.edu}\\
$^{\dagger\dagger}$\textit{Department of Health Systems \& Population Health Sciences, University of Houston}, USA\\
$^{\ddagger}$\textit{Department of Gynecologic Oncology and Reproductive Medicine, The University of Texas MD Anderson Cancer Center}, USA\\
\texttt{\{asood, SCorvigno, estur\}@mdanderson.org}%
}

\thanks{$^{*}$Corresponding author: Lu Wang (lwang71@central.uh.edu).}%
}

\maketitle
\IEEEaftertitletext{\vspace{-0.8\baselineskip}}

\begin{abstract}

Clinical decision support AI systems (CDSASs) must adapt to evolving patient conditions in real-time while adhering to strict safety constraints. We present an online adaptive framework that integrates Treatment Effect (TE) estimation to quantify clinical benefits, a patient Digital Twin (DT) to simulate treatment trajectories, and Reinforcement Learning (RL) for sequential decision-making. The AI system is initially trained on historical medical records and operates in a continuous learning loop. To ensure safety, a rule-based module monitors vital signs and blocks contraindicated treatments. Cases with strong internal model disagreement are flagged for clinician review, simulated in our experiments via a pre-trained outcome model.  We validate our framework using both a synthetic clinical simulator and a real-world ovarian cancer dataset from The Cancer Genome Atlas (TCGA). In both simulated and clinical settings, our method demonstrated superior effectiveness and stability in recommending treatments compared to standard computational baselines. Furthermore, the AI system maintains low latency and requires expert consultation for only a minority of cases in our experimental validation, demonstrating its potential as a safe, clinician-supervised tool for personalized medicine that continuously improves through practical use.

\end{abstract}
\vspace{-5pt}
\section{Introduction}
\vspace{-5pt}
Clinical decisions arrive in sequence and involve risk \citep{SuttonBarto2018}. Policies learned offline can be effective at deployment, yet dataset shift and limited coverage reduce value as conditions evolve \citep{Levine2020OfflineRL}. Our objective is an online adaptive clinical decision support tool that learns during use while respecting safety. Treatment Effect (TE) estimation serves as the primary metric for clinical benefit, ensuring the AI system prioritizes interventions that offer the greatest evidence-based improvement for the patient under a clear counterfactual reference \citep{HernanRobins2020}. A patient Digital Twin (DT) provides a virtual environment to simulate patient responses and predict potential future health states based on real-time data \citep{Viceconti2021DigitalTwinHC}. Reinforcement Learning (RL) enables long-term treatment planning by modeling the relative value of different clinical actions over time \citep{SuttonBarto2018, chen2022relax}.

We link these parts into a single AI system focused on online learning with guardrails. First, the AI system undergoes an offline training stage using historical medical records, ensuring that its recommendations remain within the boundaries of established clinical practices \citep{Fujimoto2019BCQ}. Second, during real-time operation, the tool suggests treatments by pooling insights from multiple internal models and only requests human expert guidance when these models show high uncertainty. This uncertainty is quantified by measuring the variation in predictions across the model ensemble, providing a reliable measure of confidence for clinical decision support \citep{Lakshminarayanan2017DeepEnsembles, chen2025frog, raza2025neuromoe}. Third, the AI system performs frequent, stable updates on recent patient data to adapt to evolving conditions without reacting erratically to new information \citep{Jayaraman2024PrimerRLMedicine}. To minimize the workload for clinicians, an automated selection process identifies only the most informative and diverse cases for expert review \citep{Sener2018CoreSet}. The tool exposes lightweight controls for query threshold, stream rate, and batch size, enabling simple, rapid runtime behavior changes without requiring full retraining.

\begin{figure}[t]
  \centering
  \includegraphics[width=\linewidth]{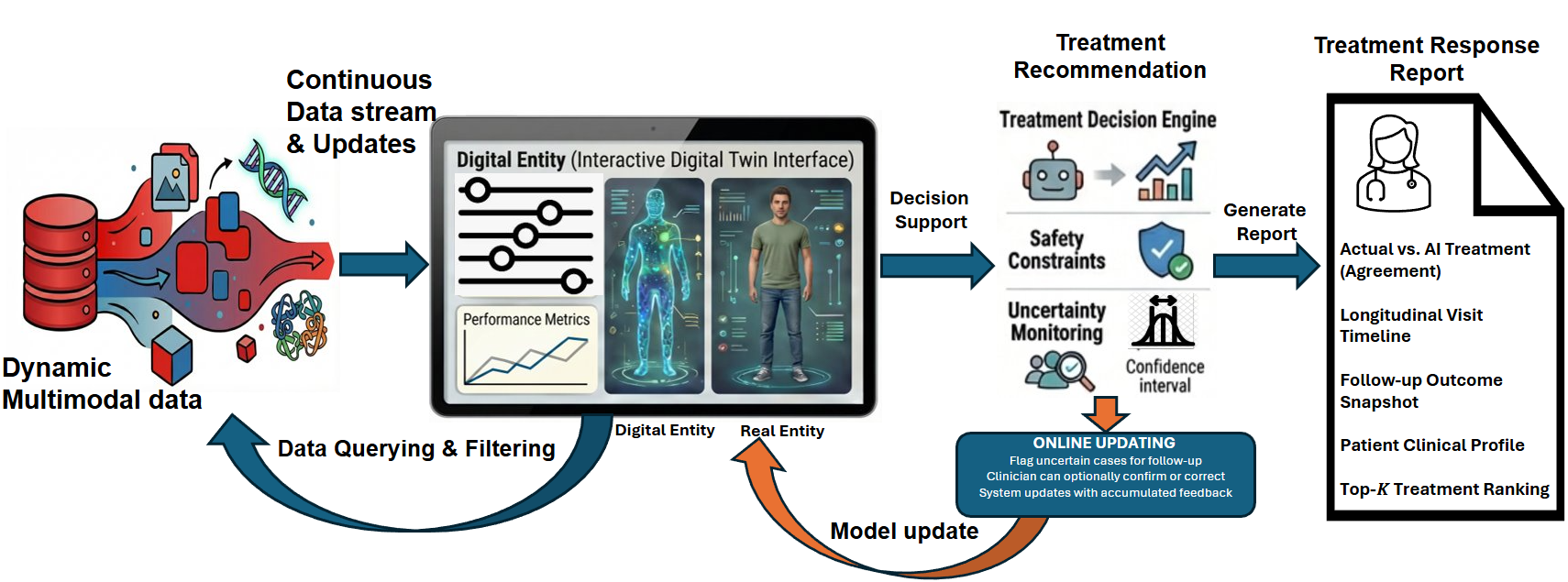}
  \caption{Overview of the proposed DT powered treatment response AI system. Dynamic multimodal data are ingested continuously to update the DT state, which supports treatment recommendation under safety constraints and uncertainty monitoring, and generates a treatment response report; an online updating loop flags uncertain cases for clinician follow-up and uses accumulated feedback for model updates.}
  \label{fig:framework}
\end{figure}

\begin{figure*}[h!] 
\centering 
\includegraphics[width=0.9\textwidth]{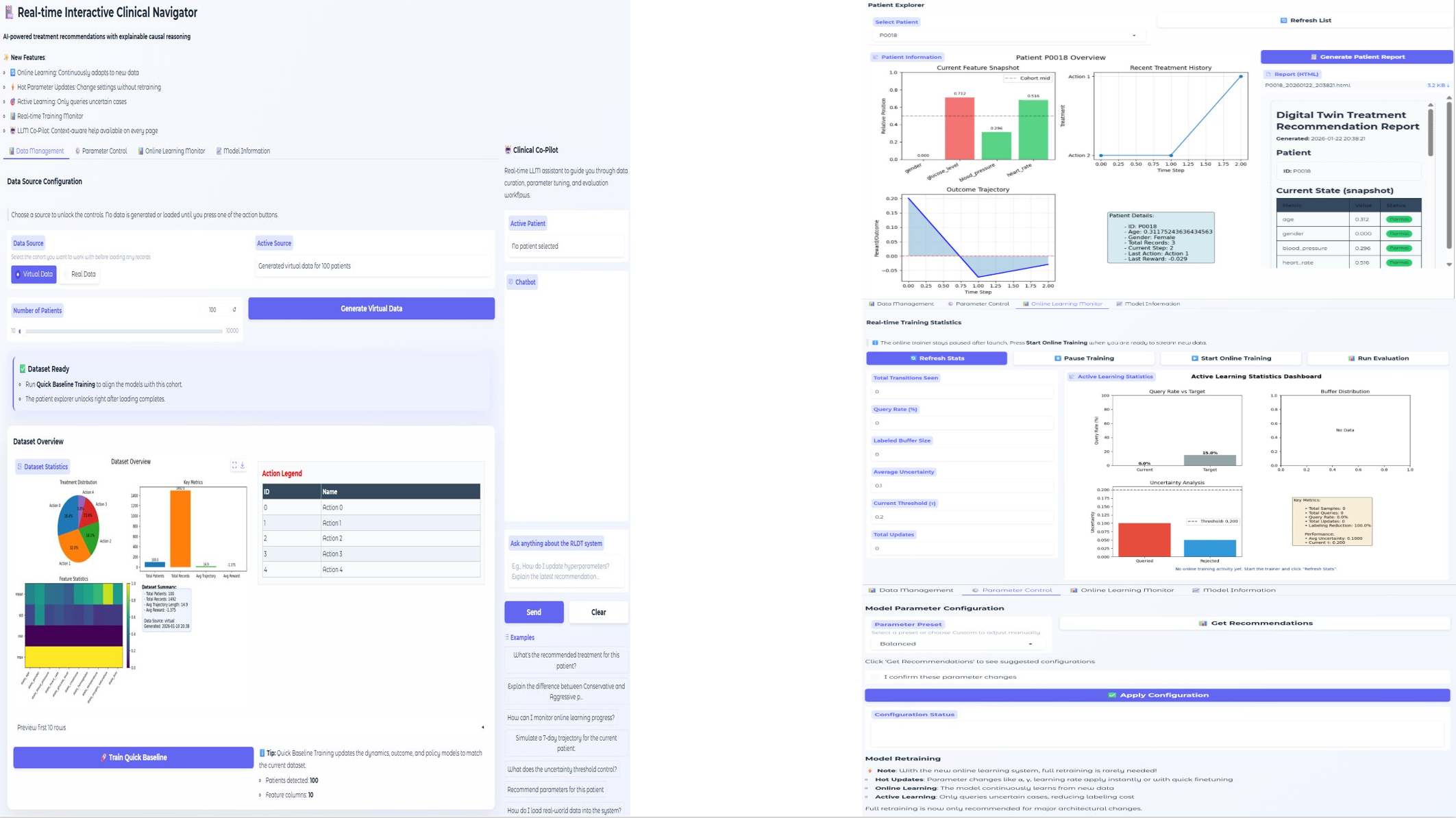} \caption{Tool overview of the deployed AI system. The left panel provides the data upload and preview interface with controls to launch model training, the right sidebar integrates a  Large Language Model (LLM)-powered AI chatbot tool for interactive assistance and AI system guidance. The top-right panel shows patient data exploration with report generation and an in-page preview (a full report example is shown in Fig. \ref{fig:case-detail}). The middle-right panel displays a monitoring dashboard for data and training status, and the bottom-right panel offers parameter controls for configuring model training and settings.} \label{fig:web-interface} 
\end{figure*}

This paper presents an online adaptive decision support framework that integrates TE estimation, DT, and RL into an online adaptive decision support tool. First, we establish a robust decision-making core by combining a stable model initialized from historical data with a responsive online update mechanism. Second, we integrate a patient DT that enables rapid and consistent health state simulations during real-time operation. Third, we guide the learning process using TE metrics, ensuring the AI system optimizes for clinically significant outcomes while maintaining safety through uncertainty monitoring and rule-based constraints. 

To validate these contributions and address the gap between simulation and clinical practice, we evaluate the framework on two datasets: a controlled synthetic environment for systematic analysis, and a real-world ovarian cancer treatment cohort derived from The Cancer Genome Atlas (TCGA) \citep{cancer2011integrated}. We selected ovarian cancer as the real-world case study because our clinical collaborator is an ovarian cancer specialist, providing a clinically grounded setting to validate our AI system's decision-making behavior and reporting workflow. The ovarian cancer dataset presents clinical challenges, including infrequent positive treatment responses (occurring in only 27.5\% of patients), complex combinations of up to 11 treatments, and detailed patient profiles encoding clinical staging and performance status. Our results demonstrate that the proposed method achieves significant improvements in both simulated and real-world clinical settings, reflecting its practical applicability. Overall, an illustration of our method is shown in Fig.~\ref{fig:framework}, and our main contributions are briefly summarized below as follows:

\begin{itemize}
    \item \textbf{Publicly deployed AI tool for immediate and broad access.} The complete implementation of our framework, encompassing all methodological details presented in this paper, is publicly deployed as an interactive web application.\footnote{\url{https://huggingface.co/spaces/KingmaoQ/RLDT}} The tool is available on-demand to any user without installation or registration. The complete AI tool overview is shown in Fig.~\ref{fig:web-interface}.
    
    \item \textbf{A safety-aware online evaluation loop for Digital Twins in Healthcare.} We integrate an uncertainty-driven query mechanism with explicit rule-based safety gates (such as vital-sign plausibility, medication dose bounds, and conflict checks) to trigger conservative fallbacks before any potential clinical violation.

    \item \textbf{Uncertainty-driven selective querying under real-time constraints.} We formalize an automated querying process that identifies informative cases by evaluating the level of agreement across multiple internal models. \citep{Lakshminarayanan2017DeepEnsembles,CheaperEnsembles2024}.    

    \item \textbf{Seamless transition from historical data to real-time adaptation.} We initialize the AI system using high-performing models trained on retrospective data and apply frequent, stable updates to balance the need for learning new patterns with the necessity of maintaining AI system stability \citep{Fujimoto2019BCQ,Jayaraman2024PrimerRLMedicine}.

    \item \textbf{Privacy-preserving data processing.} We implement a module to de-identify data at the point of entry, automatically removing direct identifiers and applying standard privacy protection techniques in compliance with the Health Insurance Portability and Accountability Act (HIPAA) Safe Harbor standards \cite{portability2012guidance}.

\end{itemize}

\vspace{-5pt}
\section{Methodology}

\vspace{-5pt}

\subsection{Offline Training with Three-Stage Model Development}

Before any model consumes data, we run a policy-driven de-identification pass so all learning uses HIPAA Safe Harbor de-identification standard. Specifically, we remove direct identifiers (such as names and medical record numbers), replace internal record identifications with random study identifications, reduce the detail of potentially identifying fields (for example, we keep only the first three digits of Zone Improvement Plan (ZIP) codes and group ages into ranges), and shift dates by a small, fixed maximum amount to prevent re-identification while preserving the relative timing of events. As an additional safeguard, we verify k-anonymity (k), meaning that each record is indistinguishable from at least k other records on the fields that could indirectly identify a person.

\subsubsection*{Stage 1: Dynamics Model (Ensemble of Five)}
We construct a patient DT that predicts the next state from recent history and the applied treatment. 
The model is a Transformer encoder that receives a sequence of state vectors and the aligned action tokens, with a causal attention mask and a padding mask \cite{lai2026transformers}. 
At each step the network predicts a residual change, and we apply a strictly bounded update to improve stability during iterative multi-step rollouts:
\vspace{-8pt}
{\small\begin{equation}
\mathbf{s}_{t+1} \;=\; \operatorname{clip}\!\Bigl(\mathbf{s}_{t} \,+\, 0.05\, \tanh\!\bigl(f_{\theta}(\mathbf{s}_{0:t}, a_{0:t})\bigr),\; 0,\; 1\Bigr).
\end{equation}}

Here $\mathbf{s}_t \in [0,1]^d$ is the normalized state and $a_t \in \{0,\dots,K-1\}$ is the discrete action. 
The loss is computed only on valid timesteps within each sequence by a binary mask that ignores padding. 
We use a Smooth L1 objective over one step predictions:
\vspace{-8pt}
{\small\begin{equation}
\mathcal{L}_{\text{DT}}(\theta) \;=\; \frac{1}{|\Omega|} \sum_{(i,t)\in \Omega} 
\ell_{\text{smooth}}\!\left(\hat{\mathbf{s}}^{(i)}_{t+1} ,\, \mathbf{s}^{(i)}_{t+1}\right),
\end{equation}}

where $\Omega$ denotes all valid positions in the mini batch. 
Training uses AdamW, gradient clipping, and a learning rate scheduler. 
We train five independent models under different seeds and keep all five for evaluation. 
During rollout we aggregate the predictions by the ensemble mean. 
We also use the ensemble variance as an uncertainty signal.


\subsubsection*{Stage 2: Counterfactual Treatment Outcome and Reward Model}
Network $r_{\phi}$ predicts the immediate outcome from $(\mathbf{s}, a)$. Let $\mathbf{z}_{\text{health}}=g_{\phi}(\mathbf{s})$ denote the health representation learned from state features. We apply adversarial deconfounding with a discriminator $D_{\xi}(a\,|\,\mathbf{z}_{\text{health}})$, where $\mathcal{L}_{\text{adv}}$ is the action-prediction cross-entropy:
\vspace{-5pt}
{\small\begin{equation}
\min_{\phi}\;\max_{\xi}\;\;\mathbb{E}_{(\mathbf{s},a,y)\sim\mathcal{D}}\!\left[\,|\,r_{\phi}(\mathbf{s},a)-y\,| 
\;+\; \lambda_{\text{adv}}\, \mathrm{CE}\!\bigl(D_{\xi}(\cdot\,|\,\mathbf{z}_{\text{health}}),\,a\bigr)\right].
\end{equation}}
Here $\mathbb{E}_{(\mathbf{s},a,y)\sim\mathcal{D}}$ denotes the expectation over the dataset $\mathcal{D}$, $|\cdot|$ is the absolute prediction error, and $\mathrm{CE}(\cdot,\cdot)$ is the cross-entropy loss for training $D_{\xi}$ to predict $a$ from $\mathbf{z}_{\text{health}}$. The weight $\lambda_{\text{adv}}>0$ balances predictive accuracy and adversarial regularization, reducing dependence on observed confounding structure in the learned representation. It does not eliminate bias from unmeasured confounders.

\subsubsection{Stage 3: Offline Policy Learning with Batch-Constrained $Q$-learning (BCQ)}
The core of our decision-making logic is centered on the concept of Quality, represented by the quality value ($Q$-value). Specifically, $Q_{\psi}(\mathbf{s}, a)$ denotes the predicted long-term clinical benefit of applying a treatment action $a$ to a patient in a given health state $\mathbf{s}$ (comprising vital signs and clinical covariates), as estimated by a neural $Q$-network with learned parameters $\psi$. To maintain clinical safety, we utilize BCQ, which limits the AI system's recommendations to a validated subset of actions $\mathcal{A}_{\text{valid}}(\mathbf{s})$. The policy $\pi(\mathbf{s})$ then identifies the optimal intervention by maximizing this quality score within the safe set:
\vspace{-5pt}
{\small\begin{equation}
\begin{aligned}
\pi(\mathbf{s}) \;&=\; \arg\max_{a \in \mathcal{A}_{\text{valid}}(\mathbf{s})} Q_{\psi}(\mathbf{s},a), \\[6pt]
\mathcal{A}_{\text{valid}}(\mathbf{s}) \;&=\; \{\,a \in \mathcal{A}\;:\; b(a\,|\,\mathbf{s}) \ge \tau_{\text{supp}}\,\}.
\end{aligned}
\end{equation}}
Within this framework, $b(a\,|\,\mathbf{s})$ represents the behavior model, which characterizes the likelihood that a human expert would select action $a$ from the total set of possible interventions $\mathcal{A}$ for a patient in state $\mathbf{s}$. The support threshold, $\tau_{\text{supp}}$, acts as a safety gate to exclude actions that lack sufficient evidence in historical clinical data, ensuring the model avoids unproven or potentially hazardous decisions. This threshold is tuned during validation to effectively balance the optimization of treatment outcomes with rigorous safety requirements.

\subsection{Online Learning with Uncertainty-Based Sampling}

High-uncertainty candidates (\( \tilde{u}(s_t)>\tau_{\text{query}} \)) are buffered; once the buffer reaches \(k\) items (query batch size), apply \(k\)-center selection (uncertainty-weighted farthest-first) to query a batch of diverse samples.

\subsubsection{ Uncertainty-Based Selective Querying}
We maintain a \emph{$Q$-ensemble} of $H=5$ independently trained $Q$-networks ${Q_{\psi_k}}_{k=1}^{H}$ initialized from the offline stage. Greedy action selection uses the ensemble mean:
\vspace{-5pt}
{\small\begin{equation}
a_t \;=\; \arg\max_{a \in \mathcal{A}} \; \frac{1}{H}\sum_{k=1}^{H} Q_{\psi_k}(s_t,a).
\end{equation}}
Ensemble mean and standard deviation:
\vspace{-5pt}
{\small\begin{equation}
\mu_a(s_t) \;=\; \frac{1}{H}\sum_{k=1}^{H} Q_{\psi_k}(s_t,a),
\end{equation}
\vspace{-5pt}
\begin{equation}
\sigma_a(s_t) \;=\; \sqrt{\frac{1}{H-1}\sum_{k=1}^{H}\!\bigl(Q_{\psi_k}(s_t,a)-\mu_a(s_t)\bigr)^2}.
\end{equation}}
Coefficient of variation:
\vspace{-8pt}
{\small\begin{equation}
\mathrm{CV}_a(s_t) \;=\; \frac{\sigma_a(s_t)}{|\mu_a(s_t)|+\epsilon}, \quad \epsilon=10^{-8}.
\end{equation}}
$\tanh$-squashed uncertainty statistic:
\vspace{-6pt}
{\small\begin{equation}
\tilde{u}(s_t) \;=\; \tanh\!\Bigl(\max_{a \in \mathcal{A}} \mathrm{CV}_a(s_t)\Bigr).
\label{eq:uncertainty-stat}
\end{equation}}

Query expert if $\tilde{u}(s_t) > \tau$ (default $\tau=0.2$). For BCQ without ensemble, use $\hat{u}(s_t) = \text{Var}(\mathbf{s}_t) / \max_{s \in \mathcal{D}} \text{Var}(s) \in [0,1]$ with same threshold. The AI system supports two modes: \emph{manual} mode where clinicians provide treatment labels and expected outcomes through the web interface, and \emph{automatic} mode where the pre-trained outcome model generates reward labels for queried transitions. In our experimental validation, we use automatic mode to simulate expert feedback.

K-center selection for batch size $k$. Let $\mathcal{U}$ be candidates exceeding threshold, $d(\cdot,\cdot)$ Euclidean distance:
\vspace{-5pt}
{\small\begin{equation}
\operatorname*{selected}
\;=\;
\arg\max_{\mathcal{B}\subseteq\mathcal{U},\,|\mathcal{B}|=k}
\;\min_{\,\mathbf{s}\in \mathcal{U}\setminus \mathcal{B}}
\;\max_{\,\mathbf{s}'\in \mathcal{B}}
d(\mathbf{s},\mathbf{s}') \cdot \tilde{u}(\mathbf{s}).
\end{equation}}
\vspace{-8pt}
\subsubsection{Incremental Model Updates}
For Transformer $\hat{f}_\theta$ with layers $\{l_1, ..., l_n\}$, freeze $\theta_{1:n-2}$, update only $\theta_{n-1:n}$:
\vspace{-6pt}
{\small\begin{equation}
\theta_{t+1}^{(n-1:n)} \;=\; 
\theta_t^{(n-1:n)} \;-\; \eta \nabla_{\theta_{n-1:n}} 
\mathcal{L}(\theta_t;\, \mathcal{D}_t^{\text{new}}).
\end{equation}}
Exponential moving average for stability:
\vspace{-6pt}
{\small\begin{equation}
\bar{\theta}_{t+1} \;=\; \alpha \bar{\theta}_t + (1-\alpha)\theta_{t+1}, 
\quad \alpha = 0.99.
\end{equation}}

\subsubsection{Experience Replay with Prioritization}
Labeled buffer $\mathcal{B}_L$ (10K) for expert-validated transitions; weak buffer $\mathcal{B}_W$ (50K) for model predictions. Prioritized sampling:
\vspace{-5pt}
{\small\begin{equation}
p(\tau_i) \;\propto\; \omega_i \cdot \exp\!\bigl(-\lambda_t \cdot (t - t_i)\bigr),
\end{equation}}
where $\omega_i$ is uncertainty weight, $\lambda_t$ controls temporal decay.
\vspace{-5pt}
\subsection{Hot Parameter Adaptation}
\vspace{-5pt}
Three-tier adaptation without full retraining. \textbf{Tier 1} (instant): uncertainty threshold $\tau$, batch size $B$, stream rate $r$, candidate actions $N$, perturbation bound $\Phi$. \textbf{Tier 2} (fast fine-tune, $M=500$ steps): discount $\gamma$, target exponential moving average (EMA) $\rho$, regularization $\lambda_{\text{reg}}$, imitation balance $\beta$; recompute targets $y_t = r_t + \gamma \max_{a'\in\mathcal{A}_N(s_{t+1})}\min_{j\in\{1,2\}} Q_{\theta_j^-}(s_{t+1},a')$ on recent data. \textbf{Tier 3} (full retrain): architecture changes, action generator changes, feature space changes, major distribution shift, or substantial drift.

\vspace{-5pt}
\subsection{LLM Integration and Clinical Interface}
\vspace{-5pt}
\subsubsection{LLM-Based Interpretability}
Tool-augmented LLM approach for natural-language explanations. LLM accesses functions (optimal action retrieval, trajectory simulation, feature importance) to generate rationales. Local LLM server \cite{openai_api,chen2026cure, cao2026task} with constraints: $<1200$ words, cite tool outputs, no hallucinated data.

\subsubsection{Human-Computer Interface and Report Generation}
Progressive disclosure interface: patient dashboard (vital signs with abnormality flags), treatment comparison panel (side-by-side projections), training monitor (live metrics). Three modes: consultation (natural-language queries), configuration (parameter tuning), monitoring (AI system performance).

Auto-generated Hypertext Markup Language (HTML) report: (1) patient profile with flagged vitals, (2) primary recommendation with confidence and expected outcome, (3) treatment comparison table, (4) decision rationale emphasizing abnormal metrics and contrasting alternatives, (5) trajectory visualizations of simulated biomarker evolution.

\vspace{-5pt}
\section{Experiments and Results}
\vspace{-5pt}
\subsection{Evaluation Datasets}
\vspace{-5pt}
We evaluate on two complementary datasets: a controlled synthetic simulator for systematic analysis and a real-world ovarian cancer cohort for clinical validation.

\subsubsection{Synthetic Clinical Simulator}
10-dimensional state space with physiologically relevant features: blood pressure (BP $\sim \mathcal{N}(0.5, 0.15^2)$), heart rate (HR $\sim \mathcal{N}(0.5, 0.1^2)$), glucose ($\sim \mathcal{N}(0.5, 0.2^2)$), creatinine, hemoglobin, temperature, oxygen saturation (SpO\textsubscript{2}), age, gender, and Body Mass Index (BMI), all normalized to $[0,1]$. Action space: $K=5$ treatments. Reward structure incentivizes normal vital signs (SpO\textsubscript{2}$>0.9$ yields bonus) while penalizing abnormal values; SpO\textsubscript{2}$<0.80$ triggers conservative fallback and mandatory expert query. Dataset: \textbf{10,000 trajectories} (max horizon 50), split \textbf{8,000/1,000/1,000} (train/val/test) by patient ID.

\subsubsection{Real-World Ovarian Cancer Dataset}
TCGA Ovarian Cancer cohort with \textbf{587 patients}, \textbf{2,552 treatment events} \cite{cancer2011integrated}. Drug names normalized into 11 therapeutic classes: platinum, taxanes, anthracyclines, antimetabolites, topoisomerase inhibitors, alkylating agents, anti-angiogenics, hormonal agents, vinca alkaloids, radiation, and other, yielding \textbf{$K=47$ treatment combinations} via multi-hot encoding. State representation includes age, gender, tumor status, grade, stage, cumulative drug count, radiation history, Eastern Cooperative Oncology Group (ECOG) performance status, Karnofsky performance score, and reserved features. Binary reward: 1 if tumor-free transition, 0 otherwise (\textbf{27.5\% positive observed}). Split: \textbf{469/59/59 patients} (train/val/test, 80/10/10). Most frequent regimens: platinum-taxane (32\%), platinum-only (27\%), triplet combinations (11\%).

\vspace{-5pt}
\subsection{Comparative Methods}
\vspace{-5pt}
Five methods evaluated with unified preprocessing, $\gamma=0.99$, five random seeds: Deep Q-Network (DQN)~\cite{Mnih2015DQN}, Double Deep Q-Network (Double DQN)~\cite{VanHasselt2016DoubleDQN}, Neural Fitted Q-Iteration (NFQ)~\cite{Riedmiller2005NFQ}, Conservative Q-Learning (CQL)~\cite{Kumar2020CQL}, and our BCQ-based approach. All methods trained on historical data using Transformer-based dynamics ensembles and treatment outcome model with adversarial deconfounding. Treatment strategies are evaluated via rollout in the learned DT environment, reporting discounted cumulative benefit. \textbf{Safety monitoring}: all recommendations verified against rule-based clinical constraints. For synthetic data, vital sign ranges (BP$\in[0.3,0.8]$, HR$\in[0.4,0.7]$, glucose$\in[0.3,0.7]$, SpO\textsubscript{2}$\in[0.85,1.0]$, temperature$\in[0.45,0.55]$) and drug contraindication checks. For ovarian cancer data, safety constraints include stage validity (Stage I--IV only), tumor status validity (tumor-free/with-tumor only), and a conservative eligibility gate requiring ECOG $\leq 2$ and age $\in[18,90]$ strictly.

\vspace{-5pt}
\subsection{Offline Evaluation and Analysis}
\label{sec:offline-eval}
\vspace{-5pt}
All methods are evaluated on held-out test data by rolling out learned policies in the DT environment and measuring cumulative treatment benefit, defined as the total reward accumulated over the rollout with future rewards discounted, together with selection consistency.

\paragraph{Treatment Selection Performance}

Table~\ref{tab:compact-results} summarizes offline and online results. On the synthetic simulator, our method achieves mean return 37.73, a 2.8\% improvement over Double DQN ($p=0.02$) with higher consistency across patient cases (Sharpe-like index 3.43 vs. 3.17). On the ovarian cancer dataset, our method demonstrates substantially better treatment selection with 136\% higher predicted benefit (33.26 vs. 14.06, $p<0.001$). This advantage reflects the AI system's ability to identify appropriate treatment combinations in sparse-reward scenarios where only 27.5\% of treatment events lead to tumor-free outcomes. Our method also produces more consistent recommendations (action entropy 0.96 vs. DQN's 1.58), reliably suggesting the same treatment for similar patients rather than alternating between options, which is critical for maintaining physician confidence. All methods maintained perfect safety (100\% compliance with vital sign constraints and contraindication checks).

\begin{table}[h]
\centering
\caption{Comparative performance across offline and online evaluation. Mean return measures cumulative treatment benefit; Query rate indicates overall expert consultation frequency. Statistical significance: $^*p<0.05$, $^{**}p<0.01$ relative to our baseline method.}
\label{tab:compact-results}
\scriptsize
\setlength{\tabcolsep}{3pt}
\renewcommand{\arraystretch}{0.95}
\begin{tabular}{@{}l|cc|cc@{}}
\toprule
\multirow{2}{*}{Algorithm} & \multicolumn{2}{c|}{\textbf{Offline (Return)}} & \multicolumn{2}{c}{\textbf{Online (Query)}} \\
\cmidrule(lr){2-3} \cmidrule(lr){4-5}
 & Syn. & Ova. & Syn. & Ova. \\
\midrule
Ours (BCQ)    & \textbf{37.73} & \textbf{33.26} & \textbf{0.131} & \textbf{0.399} \\
DQN           & 36.70$^*$      & 1.58$^{**}$    & 0.155          & 0.451 \\
Double DQN    & 36.71$^*$      & 14.06$^{**}$   & 0.137          & 0.427 \\
NFQ           & 37.51          & 7.03$^{**}$    & 0.145          & 0.574 \\
CQL           & 16.26$^{**}$   & 1.68$^{**}$    & 0.208          & 0.412 \\
\bottomrule
\multicolumn{5}{@{}l@{}}{\footnotesize Higher return = better; Lower query = less burden.} \\
\end{tabular}
\end{table}
\vspace{-5pt}
\subsection{Online Learning Evaluation}
\label{sec:online-eval}
\vspace{-5pt}

We evaluated online adaptation by replaying test data chronologically under unified conditions. Our uncertainty-based querying achieved the lowest consultation rates: 13.1\% for synthetic cases and 39.9\% for ovarian cancer cases, representing 15.5–37.0\% reductions versus baseline methods (Table~\ref{tab:compact-results}). This efficiency stems from accurate identification of clinically ambiguous scenarios via ensemble disagreement (Eq.~\ref{eq:uncertainty-stat}). Higher query rates on ovarian data (39.9–57.4\% across methods) reflect genuine clinical uncertainty in sparse-outcome settings where treatment success is variable. All methods maintained 100\% compliance with clinical constraints throughout online operation. On the synthetic simulator, we introduced a population shift after 1000 cases (simulating an influx of older, higher-risk patients with shifted vital sign distributions). Our method adapted effectively, accumulating substantially more labeled data (1,620 samples vs. 800–1,420 for baselines) and executing more frequent updates (80 blocks vs. 39–70) while maintaining rapid decision times, sustaining treatment quality despite the demographic shift.
\vspace{-5pt}
\subsection{Model Component Evaluation}
\vspace{-5pt}
We systematically assessed each component on held-out test data. The DT ensemble achieved strong predictive accuracy ($R^2=0.82$) for forecasting patient state transitions across 500 test trajectories, with prediction error remaining controlled even at 5-step projections (mean squared error (MSE)=0.006), validating the bounded update mechanism (Eq.~1). The treatment outcome model attained $R^2=0.87$ across 7,395 treatment-outcome observations with well-calibrated uncertainty estimates (expected calibration error (ECE)=0.105), confirming reliable treatment benefit predictions on average.

\vspace{-5pt}
\section{Clinical Case Analysis and Validation}
\label{sec:clinical-validation}
\vspace{-5pt}
\subsection{Representative Patient Cases}
\label{sec:representative-cases}
\vspace{-5pt}
To assess clinical plausibility, we reviewed five representative cases from the held-out TCGA test cohort. These cases span different ages, stages, grades, and treatment patterns. Table~\ref{tab:case-studies} compares the AI system's top-ranked treatment class with the treatment recorded in the dataset. The table is intended as a qualitative illustration of plausibility and alignment with observed practice, not as a population-level accuracy analysis. Across these cases, the model did not produce clinically implausible or unsupported treatment combinations.

Across these representative cases, the recommended treatment class matched the historical clinical decision in all five instances. These cases include patients aged 42--67 years, with advanced-stage disease (Stage IIIC--IV), variable tumor grade (2--3), and diverse treatment strategies including platinum-based monotherapy, platinum--taxane combinations, alternative agents, and triplet regimens. Rather than implying optimality of the historical treatment, this concordance demonstrates that the AI system’s learned policy remains aligned with real-world oncologic practice when operating under the same observational constraints. Importantly, the model did not propose clinically implausible or unsupported treatment combinations, reinforcing the effectiveness of the behavior-constrained action space and safety gating mechanisms.

\begin{table}[h]
\centering
\caption{Five representative TCGA ovarian cancer cases used for qualitative comparison between the model recommendation and the recorded treatment.}
\label{tab:case-studies}
\scriptsize
\setlength{\tabcolsep}{3pt}        
\renewcommand{\arraystretch}{1.05} 
\begingroup
\renewcommand{\arraystretch}{0.95} 
\begin{tabular}{c p{3.2cm} c c p{2.2cm}}
\toprule
\textbf{No.} & \textbf{Patient Profile} & \textbf{Rec.} & \textbf{Actual} & \textbf{Outcome} \\
\midrule
1 & Age 50, Stage IIIC, Grade 3 & Plat+Tax   & Plat+Tax   & Tumor (5 yr) \\
2 & Age 67, Stage IIIC, Grade 3 & Plat only  & Plat only  & Free (6 yr)  \\
3 & Age 42, Stage IV, Grade 3   & Alt. agent & Alt. agent & Free (6 yr)  \\
4 & Age 43, Stage IIIC, Grade 3 & Triplet Rx & Triplet Rx & Free (8 d)   \\
5 & Age 57, Stage IV, Grade 2   & Tax only   & Tax only   & Tumor (3 mo) \\
\bottomrule
\end{tabular}
\endgroup
\vspace{2pt}
{\footnotesize\raggedright
Rec.=Recommendation; Plat=Platinum; Tax=Taxane; Alt=Alternative; Rx=Treatment; yr=years; d=days; mo=months.\par
}
\end{table}
\vspace{-5pt}
\subsection{Automated Clinical Report Generation}
\label{sec:report-generation}
\vspace{-5pt}

\begin{figure}[h]
  \centering
  \includegraphics[width=0.9\columnwidth]{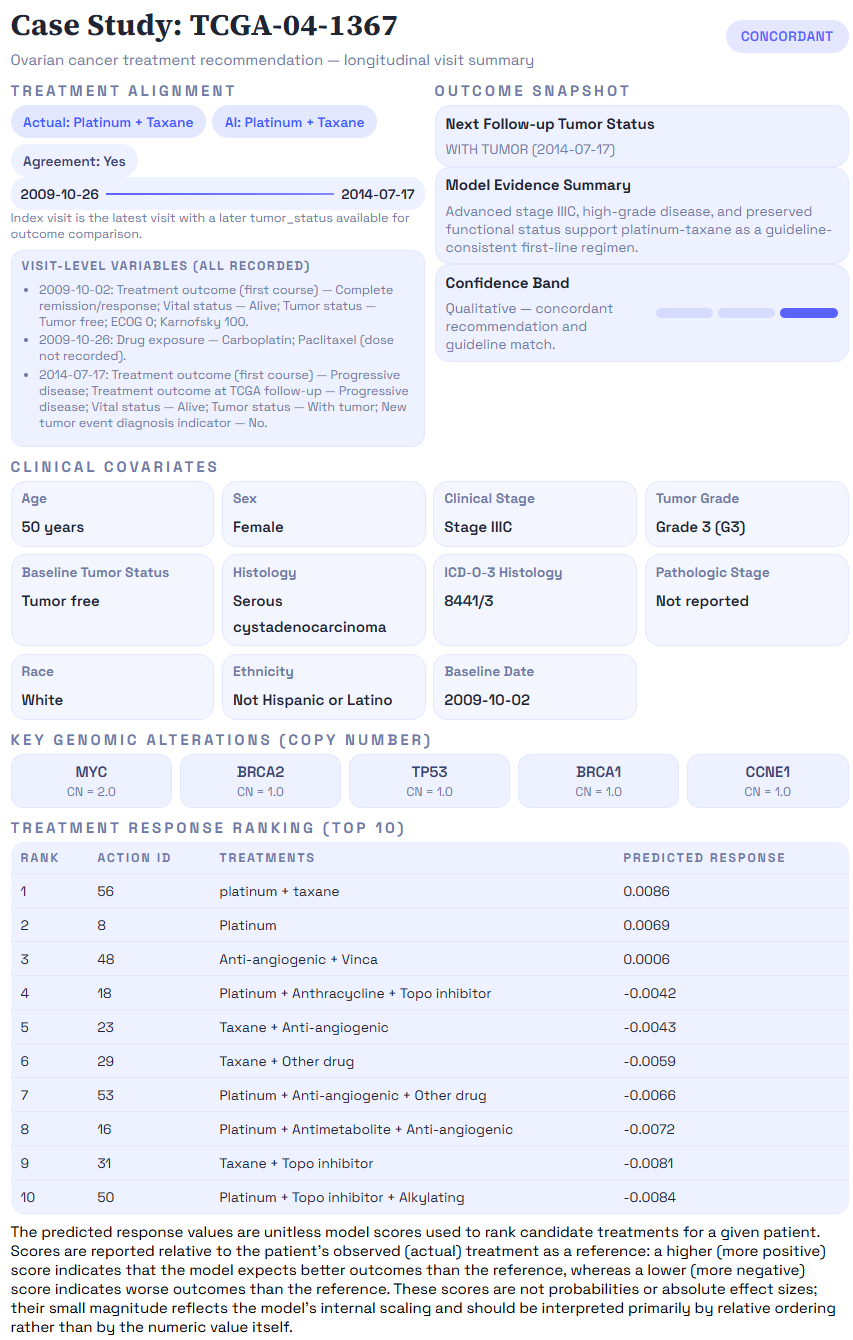}
  \caption{AI-generated patient report for TCGA-04-1367, integrating treatment rankings, clinical covariates, genomic profiles, and longitudinal outcomes.}
  \label{fig:case-detail}
\end{figure}

To illustrate how the AI system supports real-world clinical workflows, we present the automatically generated decision support report for Case \#1 (TCGA-04-1367) in Fig.~\ref{fig:case-detail}. This report demonstrates how complex model outputs, including DT simulations, TE estimates, and uncertainty metrics, are translated into a concise, clinician-oriented summary. The report ranks candidate treatment options according to predicted clinical response, where higher scores indicate greater expected benefit relative to a reference treatment. Each option corresponds to a historically observed regimen within the dataset, ensuring interpretability and clinical relevance. For this patient, the top-ranked recommendation (platinum + taxane; predicted response score = 0.0086) matched the treatment administered in clinical care.

Beyond treatment ranking, the report integrates longitudinal clinical data (2009--2014), baseline covariates, and molecular context. Genomic features include copy number (CN) status for key ovarian cancer--associated genes (MYC, BRCA1, BRCA2, TP53, CCNE1), where CN = 2.0 indicates diploid status, CN $>$ 2.0 amplification, and CN $<$ 2.0 deletion~\cite{cancer2011integrated}. These features are presented for contextualization rather than direct causal attribution, consistent with the observational nature of the dataset.

The report consolidates six integrated components: (1) \emph{Treatment Alignment}, verifying concordance between AI system recommendation and actual treatment to establish trust through demonstrated alignment with clinical practice; (2) \emph{Outcome Snapshot}, presenting tumor status trajectory with model-generated evidence summary and qualitative confidence assessment; (3) \emph{Visit-Level Variables}, documenting longitudinal treatment history and response patterns across multiple clinical encounters; (4) \emph{Clinical Covariates}, providing structured summaries of demographics, disease staging, histology, and baseline characteristics; (5) \emph{Key Genomic Alterations}, contextualizing recommendations within the patient’s molecular profile through copy number status display~\cite{cancer2011integrated}; and (6) \emph{Treatment Response Ranking}, presenting the top 10 treatment options with predicted response scores where positive values indicate expected benefit relative to reference treatment and negative values suggest potentially inferior outcomes. By integrating these elements into a unified interface, the AI system reduces the need for manual data synthesis across disparate sources and supports rapid clinical review.
\vspace{-5pt}
\subsection{Clinical Interpretation and Expert Validation}
\label{sec:expert-review}
\vspace{-5pt}
To evaluate interpretability and clinical acceptability, the generated reports are reviewed by domain experts with experience in gynecologic oncology and clinical trial design, consistent with established guidance that clinically deployed Clinical Decision Support AI System (CDSAS) should be assessed for explainability, usability, and end-user trust \cite{amann2020explainability,tonekaboni2019clinicians}. Expert feedback focused on three dimensions: (i) clinical plausibility of recommendations, (ii) clarity and completeness of the explanatory content, and (iii) perceived utility as a decision support aid rather than an autonomous decision-maker, aligning with the prevailing view that AI-enabled CDSAS is intended to support rather than supplant clinician judgment \cite{shortliffe2018cds,jones2023trust}.

Experts noted that the AI system’s recommendations were generally clinically plausible and remained within historically observed treatment classes and that uncertainty-driven escalation appropriately flagged ambiguous cases for human oversight, matching uncertainty-based referral paradigms that defer high-uncertainty cases to clinicians \cite{zhang2023calibration}. The presentation of alternative treatments with explicit comparative scores is viewed as particularly valuable for shared decision-making and hypothesis generation, especially in settings with sparse or heterogeneous evidence \cite{elwyn2012sdm}. Importantly, experts emphasized that the AI system’s primary value lies in decision support rather than replacement of clinician judgment \cite{shortliffe2018cds,jones2023trust}. The framework is perceived as most useful for synthesizing prior patient trajectories, exploring counterfactual treatment paths through DT simulation, and prioritizing options for further discussion or tumor board review. These observations align with the our AI system’s design philosophy, in which human expertise remains central and is selectively engaged when model uncertainty is high, which also helps mitigate well-described risks such as over-reliance on automated advice (automation bias) \cite{jones2023trust,goddard2012automationbias}.


\vspace{-5pt}
\section{Conclusion}
\vspace{-5pt}
We present a clinical decision-support framework that combines RL, a patient DT, and TE-based rewards with human oversight. On synthetic and retrospective ovarian cancer data, the method outperformed standard baselines while maintaining low query rates and rule-based safety. The results support the framework as a cohort-specific decision-support system within a learned DT environment. Limitations include retrospective evaluation, possible bias from unmeasured confounders, and the lack of external or prospective validation. Future work will study guideline-aware constraints, out-of-distribution detection, and multi-site prospective evaluation.

\bibliographystyle{IEEEtran}
\bibliography{refs}

@article{CheaperEnsembles2024,
  title   = {Reliable Uncertainty with Cheaper Neural Network Ensembles},
  author  = {Shui, Chang and others},
  year    = {2024},
  journal = {arXiv preprint arXiv:2403.10182},
  url     = {https://arxiv.org/abs/2403.10182}
}

@inproceedings{Kumar2020CQL,
  title     = {Conservative Q-Learning for Offline Reinforcement Learning},
  author    = {Kumar, Aviral and Zhou, Aurick and Tucker, George and Levine, Sergey},
  booktitle = {NeurIPS},
  year      = {2020},
  url       = {https://papers.nips.cc/paper/2020/hash/0d2b2061826a5df3221116a5085a6052-Abstract.html}
}

@article{Mnih2015DQN,
  title   = {Human-level control through deep reinforcement learning},
  author  = {Mnih, Volodymyr and Kavukcuoglu, Koray and et al.},
  journal = {Nature},
  year    = {2015},
  volume  = {518},
  number  = {7540},
  pages   = {529--533},
  doi     = {10.1038/nature14236}
}

@inproceedings{VanHasselt2016DoubleDQN,
  title     = {Deep Reinforcement Learning with Double Q-learning},
  author    = {van Hasselt, Hado and Guez, Arthur and Silver, David},
  booktitle = {AAAI},
  year      = {2016},
  url       = {https://arxiv.org/abs/1509.06461}
}

@inproceedings{Riedmiller2005NFQ,
  title     = {Neural Fitted Q Iteration -- First Experiences with a Data Efficient Neural Reinforcement Learning Method},
  author    = {Riedmiller, Martin},
  booktitle = {European Conference on Machine Learning (ECML)},
  year      = {2005},
  pages     = {317--328},
  publisher = {Springer}
}

@misc{openai_api,
  title        = {OpenAI API Documentation},
  author       = {{OpenAI}},
  year         = {2024},
  howpublished = {\url{https://platform.openai.com/docs/}},
  note         = {Accessed: 2025-08-18}
}

@book{SuttonBarto2018,
  title     = {Reinforcement Learning: An Introduction},
  author    = {Sutton, Richard S. and Barto, Andrew G.},
  year      = {2018},
  edition   = {2nd},
  publisher = {MIT Press}
}

@article{Jayaraman2024PrimerRLMedicine,
  title   = {A Practical Primer on Reinforcement Learning for Medicine},
  author  = {Jayaraman, Dinesh and others},
  journal = {arXiv preprint arXiv:2401.},
  year    = {2024}
}

@article{Viceconti2021DigitalTwinHC,
  title   = {Digital Twins in Healthcare: State of the Art and Challenges},
  author  = {Viceconti, Marco and Hunter, Peter and Hose, Rod},
  journal = {Annual Review of Biomedical Engineering},
  year    = {2021}
}

@book{HernanRobins2020,
  title     = {Causal Inference: What If},
  author    = {Hern{\'a}n, Miguel A. and Robins, James M.},
  year      = {2020},
  publisher = {Chapman \& Hall/CRC}
}

@inproceedings{Levine2020OfflineRL,
  title     = {Offline Reinforcement Learning: Tutorial, Review, and Perspectives},
  author    = {Levine, Sergey and Kumar, Aviral and others},
  booktitle = {NeurIPS Tutorial},
  year      = {2020}
}

@inproceedings{Fujimoto2019BCQ,
  title     = {Off-Policy Deep Reinforcement Learning without Exploration},
  author    = {Fujimoto, Scott and Meger, David and Precup, Doina},
  booktitle = {International Conference on Machine Learning (ICML)},
  year      = {2019}
}

@inproceedings{Lakshminarayanan2017DeepEnsembles,
  title     = {Simple and Scalable Predictive Uncertainty Estimation using Deep Ensembles},
  author    = {Lakshminarayanan, Balaji and Pritzel, Alexander and Blundell, Charles},
  booktitle = {Advances in Neural Information Processing Systems (NeurIPS)},
  year      = {2017}
}

@inproceedings{Sener2018CoreSet,
  title     = {Active Learning for Deep Networks: A Core-Set Approach},
  author    = {Sener, Ozan and Savarese, Silvio},
  booktitle = {International Conference on Learning Representations (ICLR)},
  year      = {2018}
}

@inproceedings{chen2022relax,
  title={Relax: Reinforcement learning agent explainer for arbitrary predictive models},
  author={Chen, Ziheng and Silvestri, Fabrizio and Wang, Jia and Zhu, He and Ahn, Hongshik and Tolomei, Gabriele},
  booktitle={Proceedings of the 31st ACM international conference on information \& knowledge management},
  pages={252--261},
  year={2022}
}

@inproceedings{lai2026transformers,
  author    = {L. Lai and Z. Cheng and K. Cheng and X. Qi},
  title     = {Do Transformers Always Win? An Empirical Study of Semantic Embeddings for Short-Text E-commerce Reviews},
  booktitle = {2026 9th International Symposium on Big Data and Applied Statistics (ISBDAS)},
  year      = {2026},
  publisher = {IEEE},
  pages     = {525--529},
  doi       = {10.1109/ISBDAS69350.2026.11484350}
}

@article{cao2026task,
  title={Task-specific efficiency analysis: When small language models outperform large language models},
  author={Cao, Jinghan and Ma, Yu and Li, Xinjin and Ren, Qingyang and Chen, Xiangyun},
  journal={arXiv preprint arXiv:2603.21389},
  year={2026}
}

@inproceedings{raza2025neuromoe,
  title={NeuroMoE: a transformer-based mixture-of-experts framework for multi-modal neurological disorder classification},
  author={Raza, Wajih Hassan and Shah, Aamir Bader and Wen, Yu and Shen, Yidan and Lemus, Juan Diego Martinez and Schiess, Mya Caryn and Ellmore, Timothy Michael and Hu, Renjie and Fu, Xin},
  booktitle={2025 47th Annual International Conference of the IEEE Engineering in Medicine and Biology Society (EMBC)},
  pages={1--7},
  year={2025},
  organization={IEEE}
}

@article{chen2026cure,
  title={CURE: Circuit-Aware Unlearning for LLM-based Recommendation},
  author={Chen, Ziheng and Cheng, Jiali and Fan, Zezhong and Amiri, Hadi and Yao, Yunzhi and Sun, Xiangguo and Zhang, Yang},
  journal={arXiv preprint arXiv:2604.04982},
  year={2026}
}

@inproceedings{chen2025frog,
  title={FROG: Fair Removal on Graph},
  author={Chen, Ziheng and Cheng, Jiali and Amiri, Hadi and Nag, Kaushiki and Lin, Lu and Liu, Sijia and Tolomei, Gabriele and Sun, Xiangguo},
  booktitle={Proceedings of the 34th ACM International Conference on Information and Knowledge Management},
  pages={415--424},
  year={2025}
}

@article{cancer2011integrated,
  title={Integrated genomic analyses of ovarian carcinoma},
  author={Cancer Genome Atlas Research Network and others},
  journal={Nature},
  volume={474},
  number={7353},
  pages={609},
  year={2011}
}

@article{amann2020explainability,
  title={Explainability for artificial intelligence in healthcare: a multidisciplinary perspective},
  author={Amann, Julia and Blasimme, Alessandro and Vayena, Effy and Frey, Dietmar and Madai, Vince I and Precise4Q Consortium},
  journal={BMC medical informatics and decision making},
  volume={20},
  number={1},
  pages={310},
  year={2020},
  publisher={Springer}
}

@inproceedings{tonekaboni2019clinicians,
  title={What clinicians want: contextualizing explainable machine learning for clinical end use},
  author={Tonekaboni, Sana and Joshi, Shalmali and McCradden, Melissa D and Goldenberg, Anna},
  booktitle={Machine learning for healthcare conference},
  pages={359--380},
  year={2019},
  organization={PMLR}
}

@article{shortliffe2018cds,
  title={Clinical decision support in the era of artificial intelligence},
  author={Shortliffe, Edward H and Sep{\'u}lveda, Martin J},
  journal={Jama},
  volume={320},
  number={21},
  pages={2199--2200},
  year={2018},
  publisher={American Medical Association}
}

@article{jones2023trust,
  title={Artificial intelligence and clinical decision support: clinicians’ perspectives on trust, trustworthiness, and liability},
  author={Jones, Caroline and Thornton, James and Wyatt, Jeremy C},
  journal={Medical law review},
  volume={31},
  number={4},
  pages={501--520},
  year={2023},
  publisher={Oxford University Press}
}

@article{zhang2023calibration,
  title={Role of calibration in uncertainty-based referral for deep learning},
  author={Zhang, Ruotao and Gatsonis, Constantine and Steingrimsson, Jon Arni},
  journal={Statistical methods in medical research},
  volume={32},
  number={5},
  pages={927--943},
  year={2023},
  publisher={SAGE Publications Sage UK: London, England}
}

@article{elwyn2012sdm,
  title={Shared decision making: a model for clinical practice},
  author={Elwyn, Glyn and Frosch, Dominick and Thomson, Richard and Joseph-Williams, Natalie and Lloyd, Amy and Kinnersley, Paul and Cording, Emma and Tomson, Dave and Dodd, Carole and Rollnick, Stephen and others},
  journal={Journal of general internal medicine},
  volume={27},
  number={10},
  pages={1361--1367},
  year={2012},
  publisher={Springer}
}

@article{goddard2012automationbias,
  title={Automation bias: a systematic review of frequency, effect mediators, and mitigators},
  author={Goddard, Kate and Roudsari, Abdul and Wyatt, Jeremy C},
  journal={Journal of the American Medical Informatics Association},
  volume={19},
  number={1},
  pages={121--127},
  year={2012},
  publisher={BMJ Group BMA House, Tavistock Square, London, WC1H 9JR}
}

@article{portability2012guidance,
  title={Guidance regarding methods for de-identification of protected health information in accordance with the health insurance portability and accountability act (HIPAA) privacy rule},
  author={Portability, Insurance and Act, Accountability},
  journal={Human Health Services: Washington, DC, USA},
  year={2012}
}

\end{document}